\documentclass[10pt,twocolumn,letterpaper]{article}

\usepackage{wacv}
\usepackage{times}
\usepackage{epsfig}
\usepackage{graphicx}
\usepackage{amsmath}
\usepackage{amssymb}
\usepackage{amsmath,amssymb}
\usepackage{multirow}
\usepackage{booktabs}
\usepackage{times}
\usepackage{epsfig}
\usepackage{graphicx}
\usepackage{algorithmic}
\usepackage{graphics}
\usepackage{epsfig}
\usepackage{amsmath,amssymb}
\usepackage{multirow}

\usepackage{enumitem}
\setenumerate{fullwidth,itemindent=\parindent,listparindent=\parindent,itemsep=0ex,partopsep=0pt,parsep=0ex}

\usepackage[belowskip=-4pt,aboveskip=0pt]{caption}

\def\wacvPaperID{535} 

\wacvfinalcopy 

\ifwacvfinal
\fi

\ifwacvfinal
\usepackage[breaklinks=true,bookmarks=false]{hyperref}
\else
\usepackage[pagebackref=true,breaklinks=true,colorlinks,bookmarks=false]{hyperref}
\fi

\ifwacvfinal
\else
\fi

\begin{document}

\title{WDNet: Watermark-Decomposition Network for Visible Watermark Removal}

\author{Yang Liu$^{1}$\thanks{\scriptsize{~Equal contribution.}}, Zhen Zhu$^{1}$\footnotemark[\value{footnote}], Xiang Bai$^1$\thanks{\scriptsize{~Corresponding author}}\\
$^1$Huazhong University of Science and Technology\\
{\tt\small \{yangl\_, zzhu, xbai\}@hust.edu.cn}
}
\maketitle
\thispagestyle{empty}
\begin{abstract}

Visible watermarks are widely-used in images to protect copyright ownership. 
Analyzing watermark removal helps to reinforce the anti-attack techniques in an adversarial way. Current removal methods normally leverage image-to-image translation techniques. Nevertheless, the uncertainty of the size, shape, color and transparency of the watermarks set a huge barrier for these methods. To combat this, we combine traditional watermarked image decomposition into a two-stage generator, called Watermark-Decomposition Network~(WDNet), where the first stage predicts a rough decomposition from the whole watermarked image and the second stage specifically centers on the watermarked area to refine the removal results. The decomposition formulation enables WDNet to separate watermarks from the images rather than simply removing them. We further show that these separated watermarks can serve as extra nutrients for building a larger training dataset and further improving removal performance. Besides, we construct a large-scale dataset named CLWD, which mainly contains colored watermarks, to fill the vacuum of colored watermark removal dataset. 
Extensive experiments on the public gray-scale dataset LVW and CLWD consistently show that the proposed WDNet outperforms the state-of-the-art approaches both in accuracy and efficiency. The code and CLWD dataset are publicly available at \url{https://github.com/MRUIL/WDNet}.

\end{abstract}

\section{Introduction}

Visible watermarks are widely used by advertisers, photographers and stock content services to mark and protect their copyrights of digital photos and videos while sharing on the Internet. Simultaneously, studying how to remove these watermarks effectively gives hints for inventing more robust techniques to better watermarking images. Considering this need, the community has put lots of efforts into the watermark removal task. 

\begin{figure*}[ht]
\centering
\includegraphics[width=\linewidth]{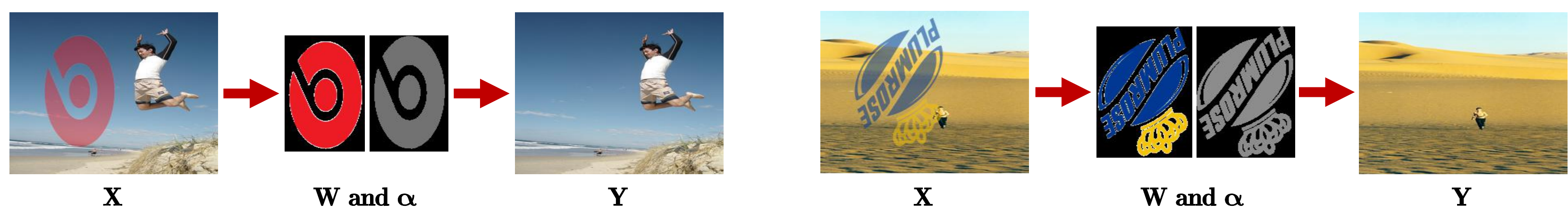}
\caption{We show the visible watermarks can be removed and separated automatically. With a single watermarked image $X$ input, our method estimates the watermark $W$ and transparency $\alpha$ first. Finally we can obtain the watermark-free image $Y$ from $W$ and $\alpha$.}
\label{fig:separation}
\centering
\end{figure*}

Early watermark removal works were generally based on watermarked image composition model~\cite{Image_inpainting_2000}. That is, a watermarked image is composed of a watermark-free image and a watermark. The natural idea is to perform the inverse decomposition on watermarked images. However, estimating the correct solution is non-trivial and time-consuming. Some methods even require user guidance~\cite{Attacking_visible_watermarking_2004, A_novel_image_recovery_2006} or multiple images as input~\cite{Dekel_2017_CVPR} for the same watermark, which becomes too strict for practical usage.

A recently published large-scale dataset LVW~\cite{Cheng2018LargeScaleVW} has driven a heat of using deep learning techniques to this task. Some works~\cite{CaoNZW19a,Li2019TowardsPV} treated watermark removal as an image-to-image translation task and used generative adversarial networks~\cite{gan_2014} to directly map watermarked images to watermark-free ones. Benefiting from the strong ability of GANs~\cite{gan_2014} to perform image translation, these methods achieved excellent removal performance. But a shortage of these methods is the incapability of watermark separation from input images, as compared to traditional methods. Separating watermarks from input images makes it easier to utilize watermarks and then helps to reinforce the anti-attack of more advanced removal method. 
Besides, another noticeable merit is the possibility to augment the training data by applying those separated watermarks to watermark-free images, which equips the model with better generalization ability for various watermarks and hence results in a boosted inference performance.

Considering the characteristics of these two kinds of methods, we are motivated to build the watermarked-image composition mechanism inside a neural generator, which would take advantages of both yet avoid their drawbacks. Our generator, called \emph{Watermark-Decomposition Network} (abbreviated as WDNet), has a huge capacity to learn from large-scale datasets, which traditional methods lack, and the ability to separate watermarks from the input images, which current deep learning based methods don't have.

One noticeable difference between WDNet and previous methods is WDNet adopts two-stage refinement strategy. The task of watermark removal implicitly contains a preliminary step: \emph{watermark region localization}, which is ignored by previous methods. However, this step \emph{explicitly} exists in WDNet, which makes WDNet naturally a two-stage generator.
In the first stage, instead of directly estimating the watermark-free images, WDNet aims to estimate the rough area of watermarks by predicting the rough watermarks and their transparencies. Then, we follow the inverse formula of creating watermarked image to determine the preliminary watermark-free image, as exemplified in Fig.~\ref{fig:separation}. We adopt U-Net~\cite{U-net_2015} as the backbone network of the first stage to efficiently and effectively forward the encoded features directly to the decoding phase.
After attaining the initial decomposed watermark-free images, the second stage of WDNet aims to refine these images by receiving extra supervision from the output. 
Note the second stage requires the network to focus on pixel-wise refinement rather than a large and redundant context. To this end, we tail a very small network as the second stage Refinement network. It is composed of several residual blocks~\cite{ResNet_2015}, which helps to guarantee local refinement and efficiency. The two-stage networks work concordantly together and can be trained jointly. Besides, WDNet does not require any post-process procedures.

The superior performance of WDNet is closely bounded to our new dataset CLWD (Colored Large-scale Watermark Dataset). The LVW dataset only contains 80 gray-scale watermarks while CLWD contains 200 colored watermarks, which are comparatively more generalized and applicable. Experiments on both datasets consistently verify the effectiveness of WDNet in watermark removal and separation.
\section{Related Work}

Braudaway~\etal~\cite{Protecting_1996_ElectronicImage} were among the earliest to use visible watermarks in digital images. They added watermarks onto input images using an adaptive and nonlinear pixel-domain technology to identify its ownership.
Unlike adding watermarks to image, many research works~\cite{A_novel_image_recovery_2006,Attacking_visible_watermarking_2004,An_automatic_visible_watermark_removal,Dekel_2017_CVPR,Visible_watermark_removal_scheme,An_automatic_visible_watermark_detection_2017} aim to remove watermarks from watermarked image and restore the image content. Pei~\etal~\cite{A_novel_image_recovery_2006} proposed to use Independent Component Analysis (ICA) to separate the source image from the watermark. Huang~\etal~\cite{Attacking_visible_watermarking_2004} used a classic image inpainting method~\cite{Image_inpainting_2000} to ﬁll in the image regions covered by the watermark. These techniques operate on a single image, requiring a user to manually mark the watermark areas, and cannot handle large watermarked regions. 
Another typical work was proposed by Deke~\etal~\cite{Dekel_2017_CVPR}, and they presented a generalized multi-image matting algorithm that assumed multiple images have the same watermark pattern, which has great limitations in real-world scenarios where the watermarks are more likely distinct in different images, and the effect depends on the numbers of multi-image.

Recently, a published large-scale dataset LVW~\cite{Cheng2018LargeScaleVW} enables training deep networks for watermarks removal. Most of them~\cite{CaoNZW19a,Li2019TowardsPV} are based on GAN~\cite{gan_2014} because of its powerful capabilities in the field of image generation. One typical work of this type is~\cite{Li2019TowardsPV}, which proposed a new watermark removal framework enabled the
watermark removal solution to be more closed to the photo-realistic reconstruction using a patch-based discriminator conditioned on the watermarked images based on cGAN~\cite{cgan_2014}, but it ignored the process of watermarked image composition. 

In nature, watermark removal resembles several content removing tasks, such as image matting~\cite{imagematting_eccv2016}, shadow~\cite{Cunaaai20} and raindrop removal~\cite{Qian_cvpr_rain}, \emph{etc.} However, the image matting task~\cite{imagematting_eccv2016} aims to predict an alpha matte from an image and a trimap while the target of watermark removal is getting watermark-free image from its corresponding watermarked image. Shadow usually appears black while watermarks can be very colorful. Removing raindrops requires detailed replenishing, which is non-trivial. However, raindrops exhibit repetitive patterns and shapes while the shapes and patterns of watermarks are quite diverse. These differences make watermark removal stand out as a unique and difficult case of content removing tasks. 

Since raindrops are small and appear all over the image, Qian~\etal~\cite{Qian_cvpr_rain} adapted the LSTM~\cite{lstm} structure to optimize attention step by step to remove raindrops progressively. For shadow removal task, Cun~\etal~\cite{Cunaaai20} utilized novel masks or scenes to erase the shadows and produce high-quality ghost-free images. These works both used GAN~\cite{gan_2014} for better performance. 
\begin{figure*}[ht]
\centering
\includegraphics[width=\linewidth]{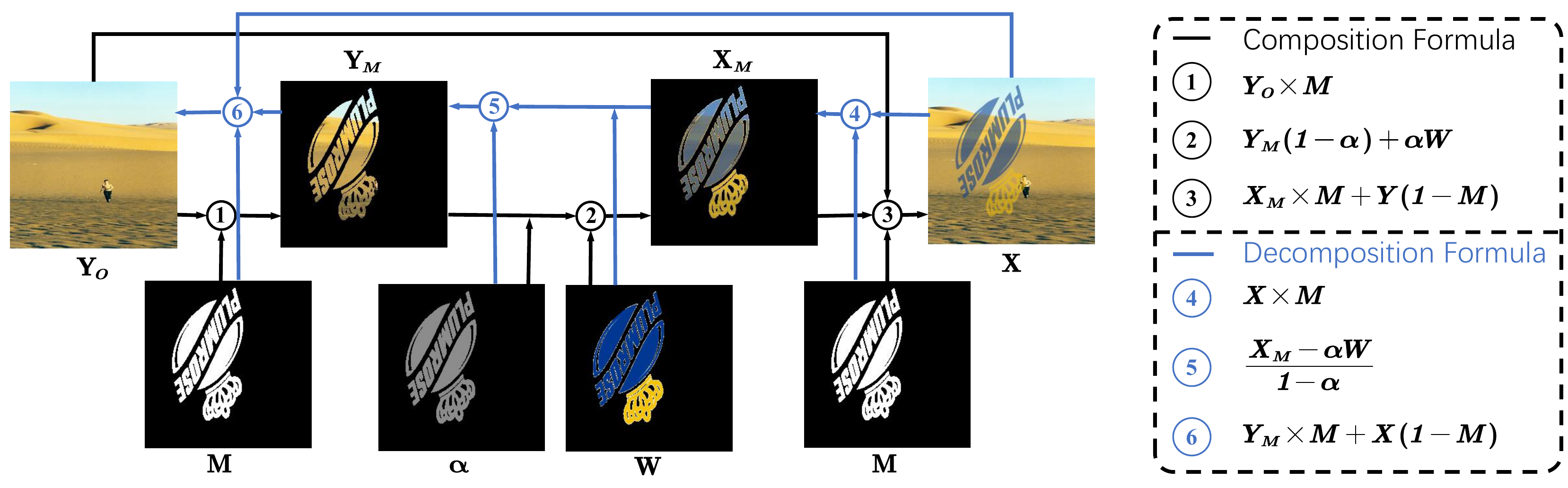}
\caption{Illustration of the watermarked image decomposition and composition model. $Y_M$ represents a masked watermark-free image, $X_M$ represents a masked watermarked image.\vspace{-1em}}
\label{fig:intro1}
\centering
\end{figure*}

Inspired by these works, our method also relies on the cGAN~\cite{cgan_2014} framework. But for good watermark removal performance, we need to consider the unique characteristics of watermarks and propose a reasonable solution. With this purpose, we analyze the composition of watermarked images and use deep networks to reflect the inverse formula of watermark composition to decompose watermarks from watermarked images. Besides, we incorporate a two-stage mechanism where the first stage is to roughly localize watermarks and the second is to refine the initial separation results. Our strategies are proved to be useful through extensive experiments.

\section{Watermarked Image Decomposition Model}
\label{sec:decomposition model}

A watermarked image $X$, is typically obtained by superimposing a watermark $W$ to a natural image $Y$. Inspired by the model of~\cite{Dekel_2017_CVPR}, in the watermarked areas, the relationship between watermarked pixels $X(p)$ and watermark-free pixels $Y(p)$ is formulated as:
\begin{equation}
\setlength{\abovedisplayskip}{3pt}
\setlength{\belowdisplayskip}{3pt}
X(p) = \alpha(p)W(p) + (1 - \alpha(p))Y(p)\label{eq:getj},
\end{equation}
Where $p = (i, j)$ represents the pixel location in the image, $\alpha(p)$ is a spatially varying opacity, namely the alpha matte used in image processing. The most commonly used watermarks are translucent to keep the underlying image content partially visible~\cite{Protecting_1996_ElectronicImage}, which means that $\alpha(p)$ is in the interval of $[0, 1]$ for all pixels. According to Eq.~\ref{eq:getj}, if $\alpha(p)=1$ everywhere, $X$ is reduced to $W$; otherwise if $\alpha(p)=0$ everywhere, $X$ is equal to $Y$.

Our task in this paper is to get a watermark-free image $Y$ from its watermarked image counterpart $X$. Considering Eq.~\ref{eq:getj}, given $W$ and $\alpha$, we could trivially invert the process of synthesizing watermarked image via the per-pixel operation:
\begin{equation}
\setlength{\abovedisplayskip}{3pt}
\setlength{\belowdisplayskip}{3pt}
Y(p) = \frac{X(p)-\alpha(p)W(p)}{1 - \alpha(p)}.\label{eq:geti}
\end{equation}

Note the non-watermarked areas are kept the same in $X$ and $Y$. When building such decomposition model inside neural networks, only considering the watermarked areas saves learning capacity and potentially is helpful for the results. Therefore, we introduce a watermark mask $M(p) \in \{0, 1\}$ to assist the decomposition phase. $M(p)=1$ means the pixel $p$ in watermarked areas. Merging $M(p)$ to Eq.~\ref{eq:geti}, the watermark-free image is obtained by:
\begin{equation}
\setlength{\abovedisplayskip}{3pt}
\setlength{\belowdisplayskip}{3pt}
Y_o(p) = M(p)\cdot Y(p) + (1 - M(p))\cdot X(p).
\label{eq:getiglobal}
\end{equation}

The watermark composition and decomposition models are shown in Fig.~\ref{fig:intro1}. 
In the next section, we illustrate how to implement decomposition model inside a neural network.

\section{Watermark Removal Framework}
\label{sec:framework}
\begin{figure*}[ht]
\centering
\includegraphics[width=\linewidth]{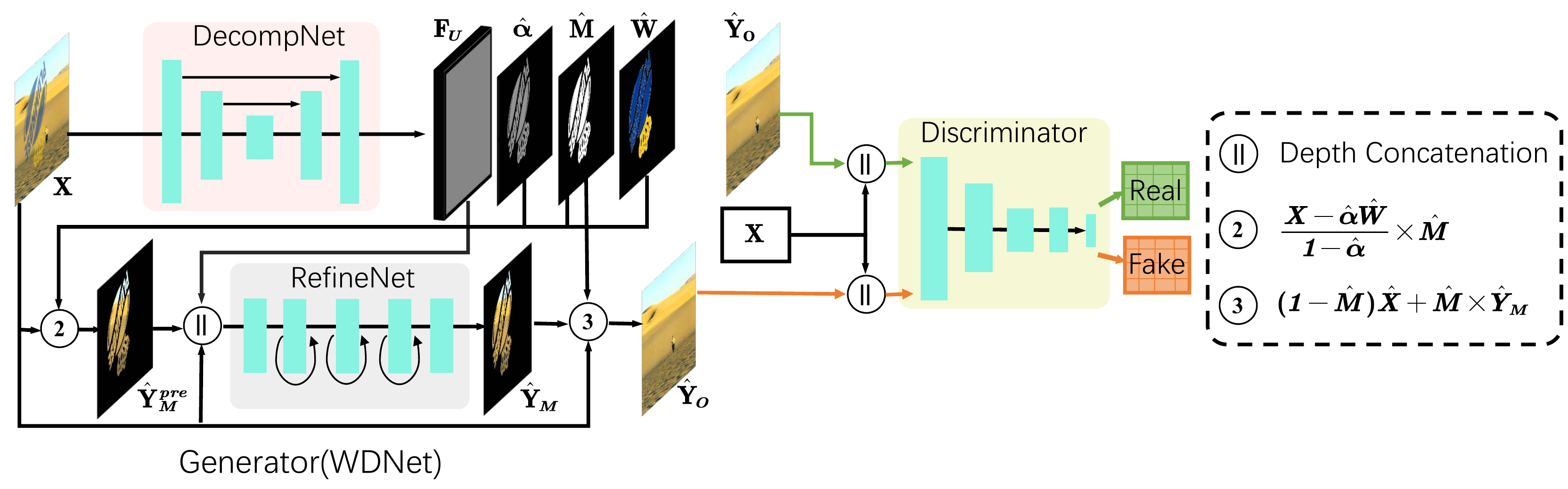}
\caption{The architecture of our visible watermark removal framework. \vspace{-1em}}
\label{fig:Generator}
\centering
\end{figure*}
Fig.~\ref{fig:Generator} sketches our framework. The whole network embodies a generator and a discriminator, where the generator, named as \textbf{Watermark-Decomposition Network}~(abbreviated as \textbf{WDNet}), utilizes the watermark decomposition model described in Sec.~\ref{sec:decomposition model} to generate the watermark-free image and the discriminator~\cite{Isola_2017_CVPR} predicts patch-wise justification scores of the image patch fidelity.

\subsection{Watermark-Decomposition Network (WDNet)}
\label{sec:WDNet}
The left part of Fig.~\ref{fig:Generator} shows the architecture of the generator WDNet, whose input is the watermarked image $X$. WDNet aims to translate the watermarked image to corresponding watermark-free image. 
Direct watermark removal from a watermarked image implicitly contains two procedures: 1) detecting the rough region of watermarked area; 2) detailed pixel-wise watermarked region denoising. From this aspect, we are inspired to devise a two-stage network where each stage aims to accomplish one of the above procedures.
Therefore, WDNet comprises of two sub-networks, which is a decomposition network in the front and a refinement network at the end. The decomposition network will initially predicts a rough result of watermark decomposition, which implies the watermarked region. The refinement network focuses on detail refinement. 

\subsubsection{The Decomposition Network}

Analyzing the watermark removal task through Eq.~\ref{eq:geti} and Eq.~\ref{eq:getiglobal},  we can calculate watermark-free image from predicting the internal $(\alpha, W, M)$, as exemplified in Fig.~\ref{fig:intro1}. However, this requires the training dataset has corresponding supervisions. Otherwise, it is much unlikely to expect good results. Thankfully, published datasets like LVW~\cite{Cheng2018LargeScaleVW} contains the $\alpha$ and $W$ for each watermarked image and we can easily calculate $M$ from $\alpha$ during data processing phase. Consequently, we first adopt a sub-network, named \textbf{DecompNet}, based on the U-Net~\cite{U-net_2015} architecture to predict the watermark parameters $(\hat{\alpha},\hat{W},\hat{M})$ from the watermarked image input. DecompNet takes advantage of the skip connections of U-Net to combine the low-level features and the high-level features together, allowing the sharing of global information and edge details between the input and the output.
Specifically, DecompNet contains $N$ down-sampling convolutional layers (the $i$-th layer with $2^{5+i}$ channels, $i \in \{1,2, \dots,N\}$) and $N$ up-sampling convolitional layers (the $j$-th layer with $2^{10-i}$ channels, $j \in \{1,2,\dots,N\}$)~($N=4$). Each feature map after the $i$-th encoding layer is bridged to the corresponding $i$-th decoding layer through depth concatenation with the previous decoded feature map. 

\subsubsection{The Refinement Network}
\label{Sec:Decomposition Net}
After the prediction of $(\hat{\alpha},\hat{W},\hat{M})$, we follow Eq.~\ref{eq:geti} and Eq.~\ref{eq:getiglobal} to calculate the preliminary watermark-free image $\hat{Y}^{pre}$.
However, the result is not satisfactory because the values of $(\hat{\alpha},\hat{W},\hat{M})$ may conflict with each other, causing unsmoothed results. Besides, this result roughly implies the watermarked area and requires further refinement. With the purpose to remedy these problems, we further bring in a small network to refine the merged output $\hat{Y}^{pre}$. The small network, named \textbf{RefineNet}, looks closely to the inferior parts in $\hat{Y}^{pre}$ and focuses on adjusting some pixels to make the final output more pleasant and smooth. 

Specifically, the first input for RefineNet is a masked preliminary watermark-free image $\hat{Y}^{pre}_M$, which is equal to $\hat{M}\cdot \hat{Y}^{pre}$. Masking out non-watermark area helps to make RefineNet avoid considering irrelevant parts and focus on refining those necessary. Another input for RefineNet is a feature map $F_U$ with 64 channels yielded from DecompNet. Feeding $F_U$ to RefineNet helps it to work closely with DecompNet when training with back-propagation algorithms. $F_U$ also provides abundant high-level semantic information, which could be substantially utilized by RefineNet. The architecture of RefineNet is quite simple with only 3 residual blocks~\cite{ResNet_2015}. Each residual block connects the previous feature map with the current one through addition and outputs a feature map with 180 channels, which helps ease training and avoid information loss~\cite{ResNet_2015}. The output of RefineNet is $\hat{Y}$. Through Eq.~\ref{eq:getiglobal}, we finally have the ultimate watermark-free image $\hat{Y}_o$, which will get judged by the discriminator later.

Through back-propagation, DecompNet and RefineNet can be trained in an end-to-end manner and provide superior watermark removal performance. Benefiting from the decomposition model and the prediction of $\hat{W}$, WDNet is armed with the ability to separate the watermarks for any watermarked images. We demonstrate in the experiments that the separated watermarks can be leveraged to further improve testing performance, which is an ideal property for practical applications.

\subsection{Discriminator}

As shown in Fig.~\ref{fig:Generator}, we employed the patch-based discriminator~\cite{Isola_2017_CVPR} in our training phase. It concatenates the watermarked image and watermark removed image as input, and maps them to a feature map, representing the probabilities of the input patches to be \emph{real} (1 for \emph{real}; 0 for \emph{fake}). Since the watermarked areas is relatively small with regards to the whole image, it will be more reasonable for the discriminator to bias towards some prominent patch statistics rather than giving a global judgement.

\subsection{Loss Function}
In summary, the loss function for the proposed method is a weighted sum of the content loss, adversarial loss, which can be seen as below:
\begin{small}
\begin{equation}
\begin{split}
\mathcal{L^{*}} =&\text{arg}\ \underset{G}{\min}\ \underset{D}{\max}\ \ \underbrace{L_{adv}(G,D)}_{\text{adversarial\ loss}} + \underbrace{L_{con}(\hat{Y_o},\hat{M},\hat{W},\hat{\alpha})}_{\text{content\ loss}}\label{eq:final_loss},
\end{split}
\end{equation}
\end{small}
\begin{table*}[]

\small
\renewcommand\arraystretch{1.2}
\begin{center}

\begin{tabular}{c|c|c|c|c|c|c|c|c}
\bottomrule[1pt] 
       Dataset & \multicolumn{4}{c|}{LVW} & \multicolumn{4}{c}{CLWD} \\ 
\bottomrule[1pt] 
\multicolumn{1}{c|}{Models} & $\mathrm{PSNR}$ & $\mathrm{SSIM}$   & $\mathrm{RMSE}$      & $\mathrm{RMSE}_{w}$      & $\mathrm{PSNR}$   & $\mathrm{SSIM}$    & $\mathrm{RMSE}$      & $\mathrm{RMSE}_w$  \\ \hline
\multicolumn{1}{c|}{Baseline}    & 37.10    & 0.9843    & 3.68   & 15.72    & 34.88      & 0.9763    & 4.71    & 25.42         \\ 
\multicolumn{1}{c|}{DecompNet}    & 41.85   & \bf{0.9966}   & 2.21    & 13.84       & 38.99   & 0.9900   & 2.72     & 19.05        \\ 
\multicolumn{1}{c|}{WDNet(Ours)}    & \bf{42.33}   & 0.9966    & \bf{2.10}    & \bf{12.38}           & \bf{40.19}   & \bf{0.9931}    & \bf{2.66}    & \bf{17.49}        \\ 
\bottomrule[1pt] 
\end{tabular}
\end{center}
\setlength{\belowcaptionskip}{-4pt}
\caption{The results of different parts of model on both the LVW and the CLWD datasets.}
\label{tab:function_analysis_3}
\end{table*}
During training, the generator $G$ is trained to minimize the adversarial objective term against the discriminator $D$, which is trained to maximize such term contrarily. Besides, the generator also tries to generate good watermark-free image, mask, watermarks and transparency, respectively. Therefore, the content loss also serves as an important part of $\mathcal{L^{*}}$.  The content loss function can be expressed as:

\begin{small}
\begin{equation}
\begin{split}
L_{con}(\hat{Y_o},\hat{M},\hat{W},\hat{\alpha}) =&\lambda_{1} L_{{1}}(\hat{Y_o}) +\lambda_{2} L_{per}(\hat{Y_o}) +\\
&\lambda_{3}L_{{1}}(\hat{M}) +\lambda_{4}(L_{{1}}(\hat{W}) + L_{{1}}(\hat{\alpha})\label{eq:cont_loss},
\end{split}
\end{equation}
\end{small}
Where $L_{{1}}(\hat{\theta})$/$L_{per}(\hat{\theta})$ refers to the $L_1$ or $L_1$ perceptual loss~\cite{PerceptualLoss_2016} between the generated object $\hat{\theta}$ and its ground truth $\theta$. $\lambda$ helps balance different loss terms. To calculate $L_1$ perceptual loss, we use the outputs of the relu2\_2 layer of a pre-trained VGG-16 network~\cite{VGG16_2014} to represent the learned features of $\hat{Y_o}$ and $Y$ and then compute their $L_1$ difference.

\section{Experiments}


\subsection{Datasets and Settings}

\subsubsection{Datasets}

As far as we are aware, publicly available datasets for watermark removal is only the Large-scale Visible Watermark Dataset (LVW)~\cite{Cheng2018LargeScaleVW}, which contains 60K watermarked images created from 80 watermarks. However, LVW mainly contains gray-scale watermarks, which is inapplicable to real cases because watermarks in the wild are often colored and distributed with various orientations. Moreover, gray-scale watermarks are monotonous and are easy to be detected, making the evaluation insufficient to test the true potential of different methods. Besides, the patterns and shapes of watermarks in LVW dataset are quite limited, bringing trouble for deep networks to capture the generalized representation of watermarks.

\begin{figure}[!t]
\centering
\includegraphics[width=\linewidth]{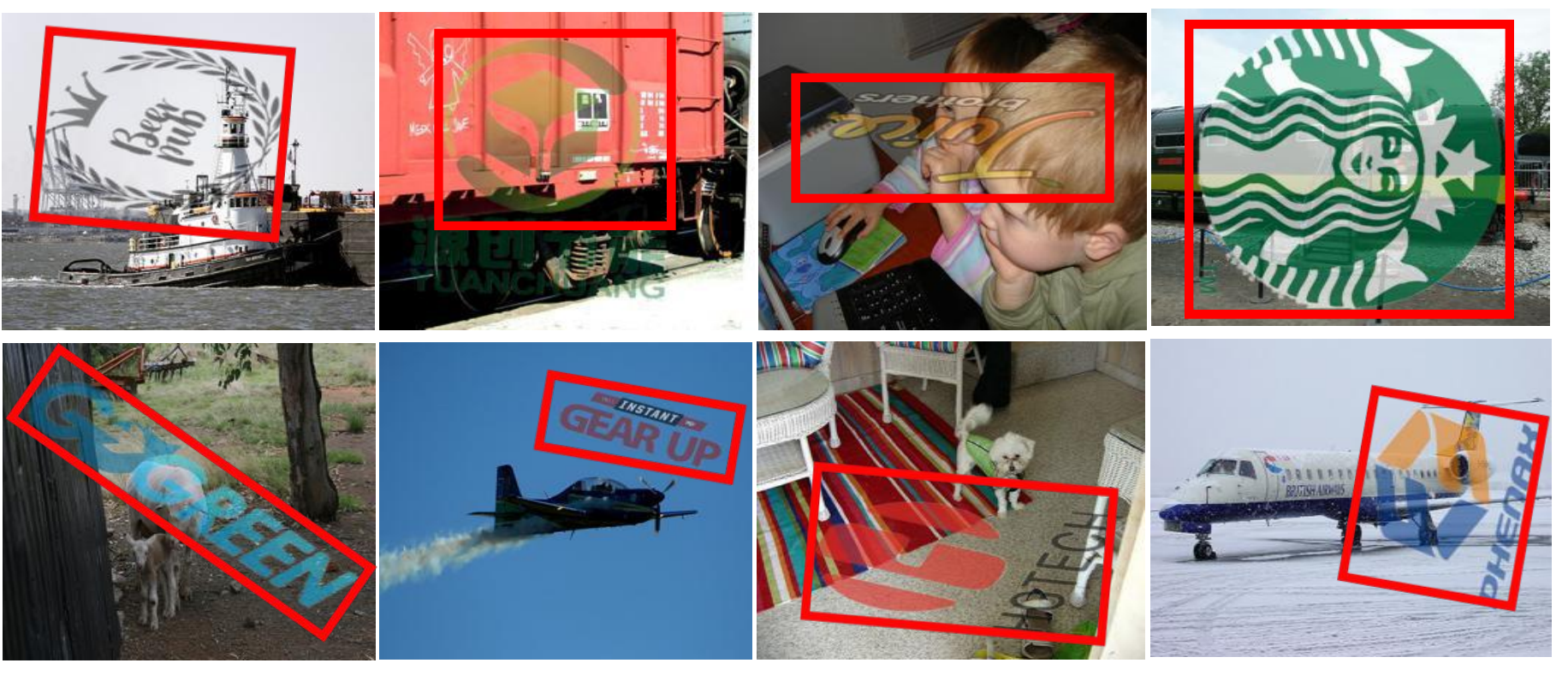}
\caption{Some examples of the CLWD dataset. }
\label{fig:CLWD}
\centering
\end{figure}

To alleviate these problems of LVW, we create a new dataset called \textbf{CLWD}~(\textbf{Colored Large-scale Watermark Dataset}), containing 60K watermarked images made of 160 colored watermarks for training and 10K watermarked images made of 40 colored watermarks for testing. 
In CLWD, the source of the watermark-free images in the training and testing sets are the PASCAL VOC2012 training and testing dataset, respectively. The watermarks are taken from the open-sourced logo images distributed online. When creating a watermarked image for training/testing, we randomly choose one PASCAL image from its training/testing set and attach one processed watermark onto it. The size, location, rotation angle and transparency of each watermark in different images are set randomly. Specifically, we set the transparency in the range of $(0.3, 0.7)$. Note CLWD also provides the corresponding watermark, mask, and transparency for each pair. Some images of CLWD dataset are shown in Fig.~\ref{fig:CLWD}. We plan to open-source the CLWD dataset for research use in the future.

\begin{figure*}[ht]
\centering
\includegraphics[width=\linewidth]{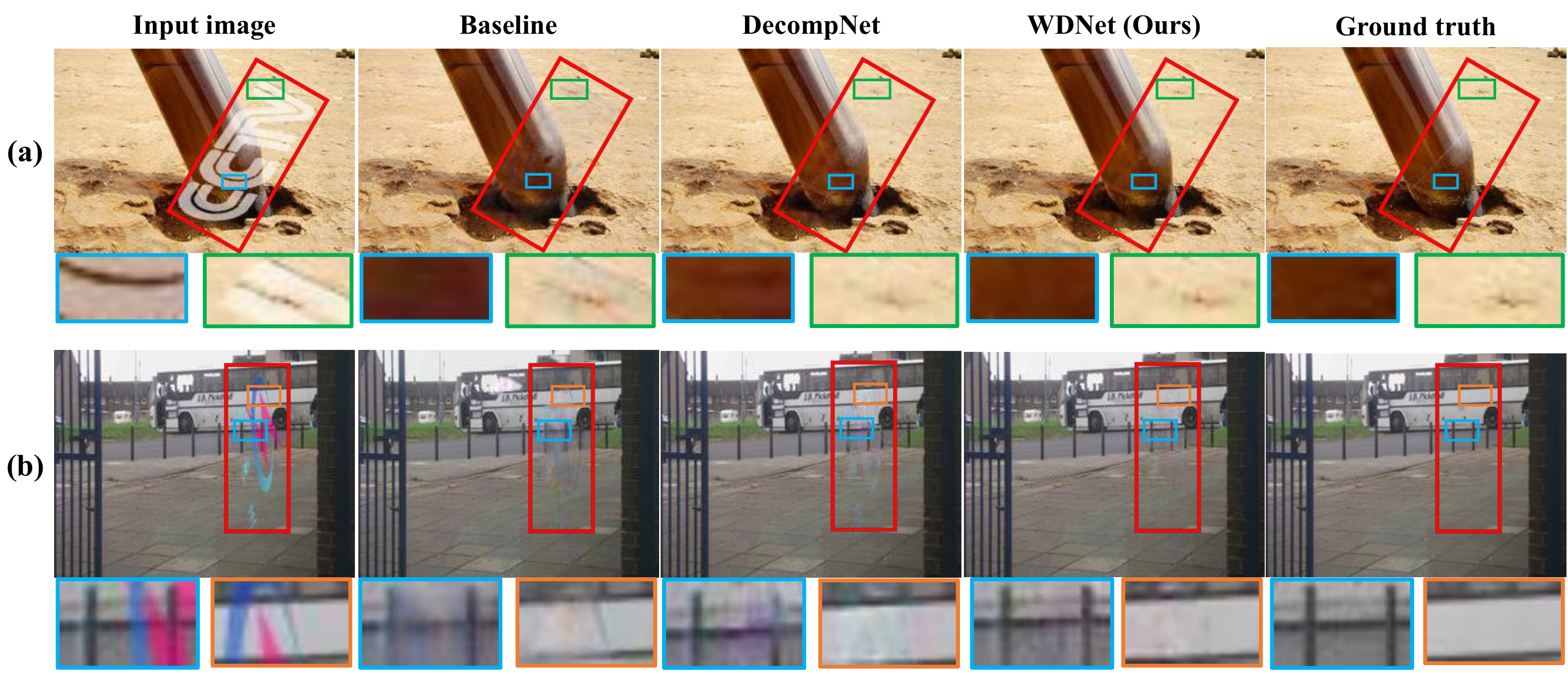}
\caption{Qualitative ablation comparisons on LVW and CLWD dataset. (a) represents the example image on LVW dataset. (b) represents the example image on CLWD dataset. \vspace{-1em}}
\label{fig:ablation_study_ab}
\centering
\end{figure*}
\subsubsection{Training details and Metrics}

\textbf{Training details.}~Our method is implemented with the PyTorch framework~\cite{Pytorch19}. All experiments are carried out on a workstation with an Intel Xeon 16-core CPU (3.5 GHz), 64GB RAM, and a single Titan Xp GPU. We follow the training strategy of the vanilla GAN~\cite{gan_2014} to alternate between one gradient descent step on the discriminator, then three steps on generator. We apply the Adam solver~\cite{adam_2014} and set batch size to 6, with the initial learning rate of 2e-4 and momentum parameters, \emph{i.e.}, $\beta_{1} = 0.5$, $\beta_{2} = 0.999$. The weights of different loss terms are $\lambda_{1} = 50$, $\lambda_{2}= 1e-2$, $\lambda_{3}=10$, $\lambda_{4}=10$, respectively. 

\noindent \textbf{Metrics.}~Following previous works~\cite{Dekel_2017_CVPR,Cheng2018LargeScaleVW}, both the Peak Signal to Noise Ratio ($\mathrm{PSNR}$) and Structural Similarity Image Index~($\mathrm{SSIM}$)~\cite{WangSSIM04}, measuring the similarity between the recovered image and the ground truth one, are adopted as our evaluation metrics. Besides, we also adopt the Root-Mean-Square ($\mathrm{RMSE}$) distance to test local similarity. In order to look at how different methods behave only in the watermarked areas, we additionally mask out the non-watermark area and then compute the Root-Mean-Square distance. This metric is denoted with $\mathrm{RMSE}_{w}$.

\subsection{Ablation Study}
\subsubsection{Different Parts of Model}

We first conduct experiments to analyze the effects of different parts of our model. Specifically, WDNet is composed by DecompNet and RefineNet and thus we design the following experiments to verify their roles to the performance:
\begin{itemize}[itemsep=2pt,topsep=0pt,listparindent=\parindent,parsep=0pt]
    \item[$\bullet$] \textbf{Baseline}~is the experiment that the generator is an U-Net and aims to generate watermark-free image from watermarked image directly without the watermark decomposition model;
    \item[$\bullet$] \textbf{DecompNet}~represents the result gained from training DecompNet separately, which embodies the watermark decomposition model;
    \item[$\bullet$] \textbf{WDNet}~represents the result of our full generator.
\end{itemize}

\begin{table}[t]
\small
\renewcommand\arraystretch{1.2}
\begin{center}
\begin{tabular}{|c|c|c|c|c|c|c|c}
\hline
Dataset      & $\mathcal{S}$  &$\mathrm{PSNR}$    &$\mathrm{SSIM}$ &$\mathrm{RMSE}$ & $\mathrm{RMSE}_{w}$      \\ \hline
LVW &              &41.55&0.9955&2.30   &14.23            \\ \hline
LVW & \checkmark    &\bf{42.33}&\bf{0.9966}   & \bf{2.10}  &  \bf{12.38}    \\ \hline
CLWD        &              &39.57 &0.9916 & 2.86  & 18.90            \\ \hline
CLWD        & \checkmark    &\bf{40.24}  &\bf{0.9931}& \bf{2.66}  & \bf{17.49}      \\ \hline
\end{tabular}
\end{center}
\caption{Effects of the $\alpha$ and $W$ supervisions($\mathcal{S}$) on LVW and CLWD datasets.}
\label{tab:midsupervision}
\end{table}
\begin{figure*}[ht]
\centering
\includegraphics[width=\linewidth]{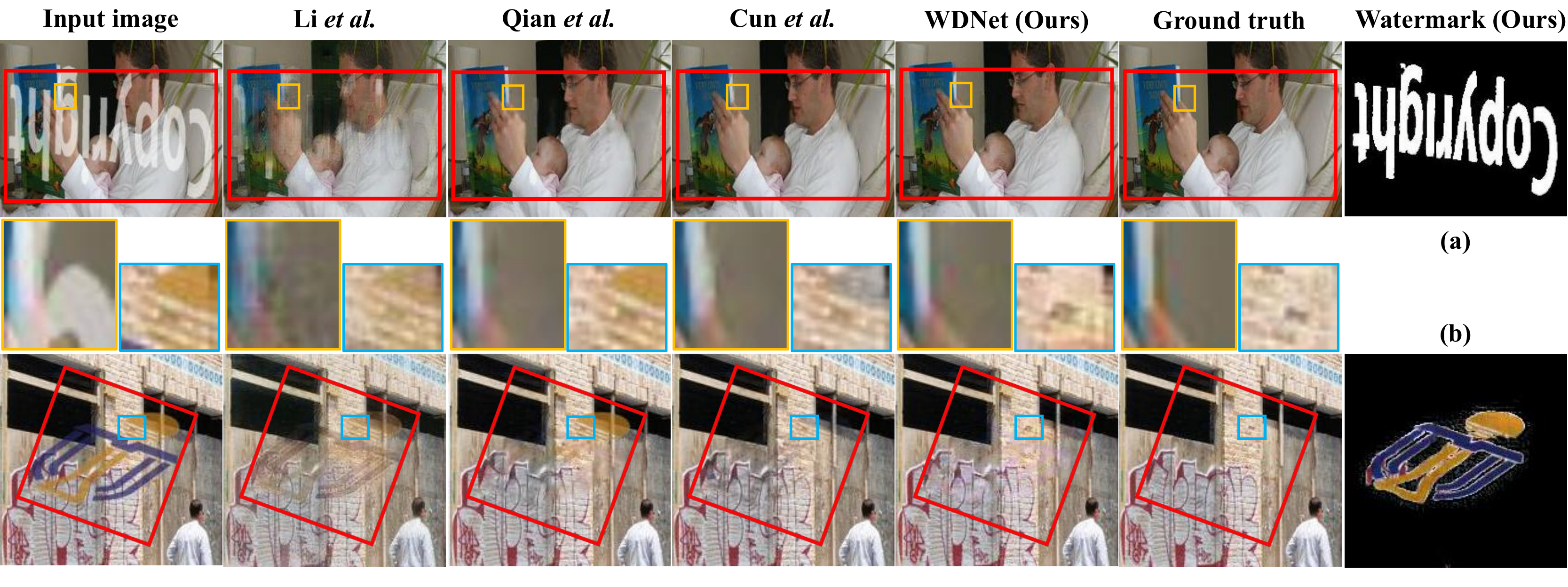}
\caption{Qualitative comparisons with some other methods on some images from LVW(a) and CLWD(b). The watermarks in the last column are obtained by our WDNet.}
\label{fig:compare_results}
\centering
\end{figure*}
\begin{table*}[]
\small
\renewcommand\arraystretch{1.2}
\begin{center}
\begin{tabular}{c|c|c|c|c|c|c|c}
\toprule[1pt]
Dataset &\multicolumn{1}{c|}{Methods} & Tasks & $\mathrm{PSNR}$ & $\mathrm{SSIM}$   & $\mathrm{RMSE}$      & $\mathrm{RMSE}_{w}$ &Time(ms)  \\ \hline
\multirow{4}{*}{LVW} &\multicolumn{1}{c|}{Qian \etal~\cite{Qian_cvpr_rain}} & Raindrop Removal &40.84  &0.9953 &2.44   &19.07 &12  \\ 
&\multicolumn{1}{c|}{Cun \etal~\cite{Cunaaai20}} &Shadow Removal &40.94  &0.9961 &2.41   &16.00 &70  \\ 
&\multicolumn{1}{c|}{Li \etal~\cite{Li2019TowardsPV}}    &Watermark Removal & 34.01 & 0.9700    & 5.34   & 26.24   &5      \\ 
&\multicolumn{1}{c|}{WDNet(Ours)}    &Watermark Removal    & \bf{42.33}   & \bf{0.9966}    & \bf{2.10}   & \bf{12.38}            &8  \\
\bottomrule[2pt]
\multirow{4}{*}{CLWD} &\multicolumn{1}{c|}{Qian \etal~\cite{Qian_cvpr_rain}} & Raindrop Removal & 39.19    & 0.9910     & 2.95   & 24.40  &16 \\
&\multicolumn{1}{c|}{Cun \etal~\cite{Cunaaai20}} &Shadow Removal  & \bf{40.32}    & 0.9899     & 2.66   & 25.59  &71 \\
&\multicolumn{1}{c|}{Li \etal~\cite{Li2019TowardsPV}}    &Watermark Removal & 32.62      & 0.9664   & 6.07     & 34.16   &6  \\
&\multicolumn{1}{c|}{WDNet(Ours)}    &Watermark Removal & 40.24   & \bf{0.9931}   & \bf{2.66}     & \bf{17.49} &8 \\
\bottomrule[1pt]
\end{tabular}
\end{center}
\caption{In-dataset evaluation results on both the LVW and the CLWD datasets.\vspace{-1em}}
\label{tab:state_of_art}
\end{table*}

The quantitative and qualitative comparison results are in Tab.~\ref{tab:function_analysis_3} and Fig.~\ref{fig:ablation_study_ab}, respectively. Comparing the performances of {Baseline} and {DecompNet}, it is evident that using the watermark decomposition model can significantly improve the watermark removal performance. Note {DecompNet} also supports watermark separation while {Baseline} doesn't have such ability.
From {DecompNet} and {WDNet}, we can learn that the RefineNet is of great help to refine watermark-free image, especially in the watermark areas, seeing a large boost of $\mathrm{RMSE}_w$ on both datasets.

\subsubsection{Intermediate Supervision}
\indent While training WDNet, we have included intermediate supervision upon the output of DecompNet. In this experiment, we aim to show the advantage of this design on both LVW and CLWD datasets.
As depicted in Tab.~\ref{tab:midsupervision}, adding intermediate supervisions of $\alpha$ and $W$ makes the model achieve improved performance over that without intermediate supervisions. It should be noted that even though without the intermediate supervision loss function, the performance of WDNet is still reasonably good, seeing that the performance drop is not significant and the result is better than most comparable methods as shown in Tab.~\ref{tab:state_of_art}.


\subsection{Comparison with State-of-the-art Methods}
\begin{figure*}[ht]
\centering
\includegraphics[width=\linewidth]{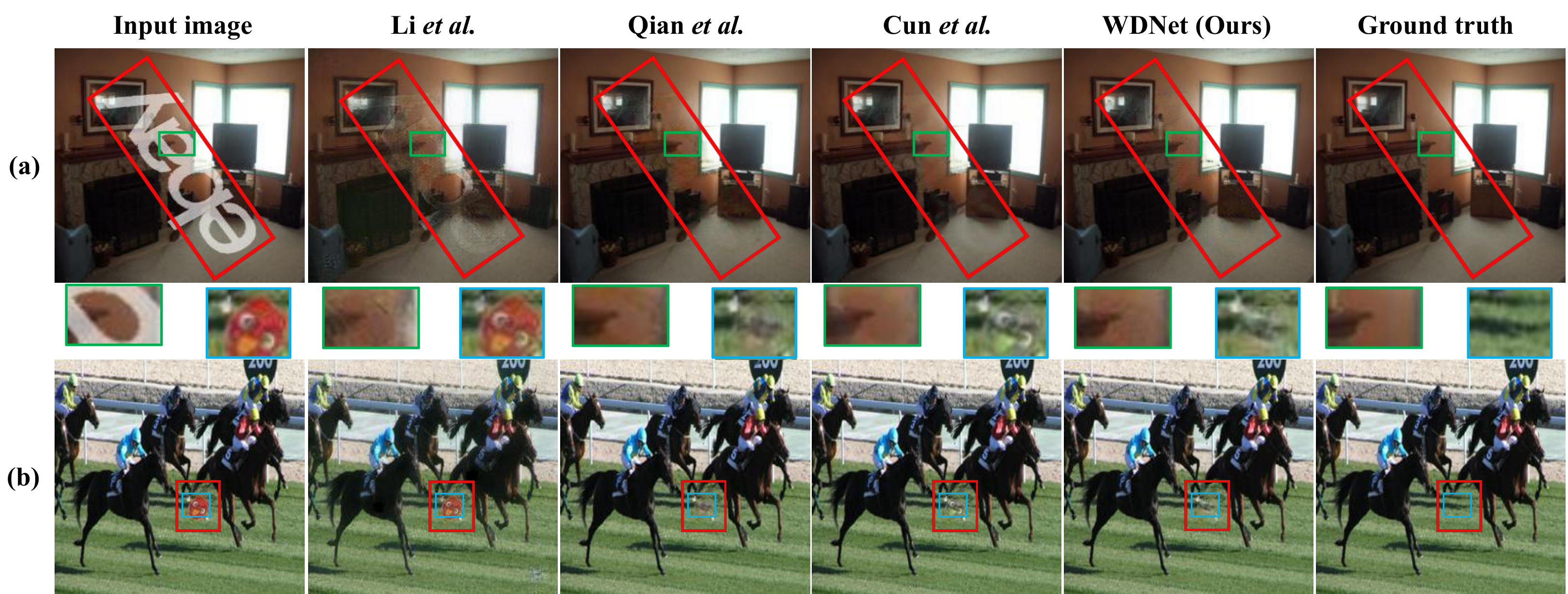}
\caption{Some failure cases in the LVW(a) and the CLWD(b) datasets. }
\label{fig:failure_case}
\setlength{\belowcaptionskip}{-3pt}
\centering
\end{figure*}

\subsubsection{In-Dataset Evaluation}


\begin{table*}[]
\small
\renewcommand\arraystretch{1.2}
\begin{center}
\begin{tabular}{cc|c|c|c|c|c|c|c}
\toprule[1pt]
        & \multicolumn{4}{c|}{CLWD(train) $\rightarrow$ LVW(test)} & \multicolumn{4}{c}{LVW(train) $\rightarrow$ CLWD(test)} \\ \hline
\multicolumn{1}{c|}{Methods} & $\mathrm{PSNR}$ & $\mathrm{SSIM}$   & $\mathrm{RMSE}$      & $\mathrm{RMSE}_{w}$      & $\mathrm{PSNR}$   & $\mathrm{SSIM}$    & $\mathrm{RMSE}$      & $\mathrm{RMSE}_{w}$  \\ \hline
\multicolumn{1}{c|}{Li \etal~\cite{Li2019TowardsPV}}    & 32.71    & 0.9668       & 6.01  & 31.78   & 30.85          & 0.9554   & 6.65     & \bf{38.29}         \\ 
\multicolumn{1}{c|}{WDNet(Ours)}    & \bf{40.86}   & \bf{0.9939}   & \bf{2.46}    & \bf{16.71}            & \bf{37.75}   & \bf{0.9754}   & \bf{3.50}     & 41.27        \\ \bottomrule[1pt]
\end{tabular}
\end{center}
\caption{Cross-dataset evaluation results on both the LVW and the CLWD datasets.\vspace{0em}}
\label{tab:state_of_art_cross}
\end{table*}
\begin{table*}[!t]
\small
\renewcommand\arraystretch{1.2}
\begin{center}
\begin{tabular}{c|c|c|c|c|c|c|c}
\toprule[1pt]
Method & Dataset & \# Watermark & \# Train pairs & $\mathrm{PSNR}$ & $\mathrm{SSIM}$ & $\mathrm{RMSE}$ & $\mathrm{RMSE}_{w}$\\
\bottomrule[1pt]

 \multirow{2}{*}{WDNet} & CLWD-100  & 100 & 60K & 40.20      & 0.9921               & 2.68 & 19.93\\
& Aug. CLWD    & 160 & 90K       & \bf{40.43}      & \bf{0.9932}             & \bf{2.61} & \bf{18.69}\\ 
\bottomrule[1pt]
\end{tabular}
\end{center}

\caption{Dataset augmentation results of WDNet.\vspace{-1em}}
\label{tab:augmentation}
\end{table*}
To justify the effectiveness of the proposed WDNet, we performed experiments to compare the WDNet with other state-of-the-art methods proposed for different tasks on both LVW and CLWD. We use the original implementations of other methods and follow their original training policies to ensure better re-implementation. As shown in Tab.~\ref{tab:state_of_art}, our model obtained consistently better results than other methods on most metrics. These results well explain the superiority of WDNet in performing the water removal task. Among all competitors, our running speed is only slower than Li~\etal~\cite{Li2019TowardsPV} while still faster than other methods, which indicates our method achieves a good balance between performance and efficiency.

In Fig.~\ref{fig:compare_results}, we also give some representative watermark removal visual results on the LVW and CLWD datasets. 
Our results show the best structural consistency with ground truth images. Beyond this, WDNet is also able to separate watermarks and watermark-free images, as shown in the last column of Fig.~\ref{fig:compare_results}, which can be used to augment training dataset as will be discussed in Sec.~\ref{sec:augmentation}.

\subsubsection{Cross-Dataset Evaluation}

Good generalization ability is always a dreaming merit of a model.  In this section, we design several experiments to train a model on one dataset and test its performance on another dataset to conduct cross-dataset evaluation. Such series of experiments have two benefits: 1) under the same training and testing setting, comparison of different models can show the generalization ability comparison of different models; 2) under the same model, comparison of different training and testing settings can show the quality comparison of different datasets. We mainly compare with Li~\etal~\cite{Li2019TowardsPV}, which achieved the previous state-of-the-art performance in watermark removal task. As depicted in Tab.~\ref{tab:state_of_art_cross}, if looking vertically (in column), we could draw the conclusion that WDNet is superior to Li~\etal~\cite{Li2019TowardsPV} in generation ability and is more robust to unseen data patterns. If looking horizontally (in row), the performance discrepancy of the same model under the settings of \textbf{LVW}$\rightarrow$\textbf{CLWD} and \textbf{CLWD}$\rightarrow$\textbf{LVW} indicates that CLWD contains more data patterns and is more diverse and generalized than LVW.





\subsubsection{Dataset Augmentation via Watermark Separation}
\label{sec:augmentation}

Many deep learning works focus on learning from more obtainable, weakly-supervised, or synthetic data~\cite{VezhnevetsFB_cvpr12,KarrasALL18,LiuFDXHHY14}. As mentioned in many parts of our paper, WDNet has the ability to separate watermarks (some examples are shown in Fig.~\ref{fig:compare_results}). This ability enables us to use those separated watermarks in many different ways, such as augmenting the training data. In this part, we conduct experiments to show we further boost the testing performance when adding separated watermarks (not included in the original training data) into training data. First, we divide the CLWD into two subsets, which are CLWD-100~(containing 100 watermarks in CLWD, 60K training pairs) and unlabeled CLWD-60~(containing 60 colored watermarks in CLWD, 30K training pairs). Second, we use the trained WDNet model on CLWD-100 to separate the watermarks of the unlabeled CLWD-60. Then, we add these watermarks into the training data of CLWD-100 by creating another 30K image pairs, resulting in an augmented dataset called Aug. CLWD. The quantitative comparison of WDNet trained on \textbf{CLWD-100} and \textbf{Aug. CLWD} is given in Tab.~\ref{tab:augmentation}. We find although CLWD-100 is robust and generalized enough, WDNet still observes further testing performance enhancement through augmenting the training set with more different watermark styles. This result also reveals the applicability of watermark separation through WDNet.

\section{Conclusions}

We propose a new watermark removal method named WDNet, which utilizes the process of watermark decomposition to roughly localize and separate watermarks first, and then uses a small RefineNet to refine the previous result in detail. In addition to the state-of-the-art performance of watermark removal, WDNet also has the ability to separate watermarks from the input images. We demonstrate that the watermarks separated by WDNet from unseen watermarked images are helpful to create more data for training, and further enhance the testing performance. Moreover, to overcome the deficiencies of the LVW dataset, we created a new watermark removal dataset, CLWD, which contains mainly colored watermarks and is more suited to real-world applications. 
Tough WDNet shows superior performance, watermark removal is extremely challenging and there are cases that WDNet fails to achieve faithful results, as illustrated in Fig.~\ref{fig:failure_case}. This drives us to investigate these cases more closely in the future. And we believe the research of watermark removal would inspire more robust techniques and strategies to defense removal attack to secure better copyright ownership and it is also in our future research plans.
~\\

\noindent \textbf{Acknowledgement}~This work was supported by the National Natural Science Foundation of China Nos. 61733007, 61703049.

\newpage
{\small
\bibliographystyle{ieee_fullname}
\bibliography{egbib}
}
\end{document}


\title{WDNet: Watermark-Decomposition Network for Visible Watermark Removal}

\author{First Author\\
Institution1\\
Institution1 address\\
{\tt\small firstauthor@i1.org}
\and
Second Author\\
Institution2\\
First line of institution2 address\\
{\tt\small secondauthor@i2.org}
}

\maketitle




\input{7_supplementary}